\documentclass[letterpaper, 10 pt, conference]{ieeeconf}  

\usepackage{graphicx}
\usepackage{amsmath}
\usepackage{amssymb}

\usepackage{soul}
\usepackage{verbatim}
\usepackage{subfigure}
\usepackage{calc}
\usepackage{amstext}
\usepackage{stackengine}
\newcommand\IncG[2][]{\addstackgap{%
\raisebox{-.5\height}{\includegraphics[#1]{#2}}}}
\usepackage{array}
\usepackage{amsmath,amssymb} 
\usepackage{xcolor}
\usepackage{paralist}
\usepackage{multirow}
\usepackage{algorithm}  
\usepackage{algpseudocode}

\IEEEoverridecommandlockouts                              

\title{\LARGE \bf
A Two-stage Unsupervised Approach for Low light Image Enhancement
}

\author{Junjie Hu$^{1}$, Xiyue Guo$^{1}$, Junfeng Chen$^{1}$, Guanqi Liang$^{2}$, Fuqin Deng$^{1}$, and Tin Lun Lam$^{1,2,\dagger} $
\thanks{This paper is partially supported by funding 2019-INT008 from the Shenzhen Institute of Artificial Intelligence and Robotics for Society.}
\thanks{$^{1}$Authors are with the Shenzhen Institute of Artificial Intelligence and Robotics for Society.}%
\thanks{$^{2}$Authors are with the Chinese University of Hong Kong, Shenzhen.}%
\thanks{$^{\dagger}$Corresponding author: Tin Lun Lam
        {\tt\small tllam@cuhk.edu.cn}
        }%
 }
\begin{document}

\maketitle
\thispagestyle{empty}
\pagestyle{empty}

\begin{abstract}
As vision based perception methods are usually built on the normal light assumption, there will be a serious safety issue when deploying them into low light environments. Recently, deep learning based methods have been proposed to enhance low light images by penalizing the pixel-wise loss of low light and normal light images. However, most of them suffer from the following problems: 1) the need of pairs of low light and normal light images for training, 2) the poor performance for dark images, 3) the amplification of noise. To alleviate these problems, in this paper, we propose a two-stage unsupervised method that decomposes the low light image enhancement into a pre-enhancement and a post-refinement problem.  In the first stage, we pre-enhance a low light image with a conventional Retinex based method. In the second stage, we use a refinement network learned with adversarial training for further improvement of the image quality. The experimental results show that our method outperforms previous methods on four benchmark datasets. In addition, we show that our method can significantly improve feature points matching and simultaneous localization and mapping in low light conditions.
\end{abstract}

\section{INTRODUCTION}
In recent years, vision based algorithms have brought significant progresses for robot's perception on various tasks such as simultaneous localization and mapping (SLAM) \cite{8025618}, object recognition \cite {he2016deep}, depth estimation \cite{hu2019visualization,ma2017sparse}, and semantic segmentation \cite{nekrasov2019real,milioto2018real,milan2018semantic}, etc. However, these algorithms are built upon the assumption that images are captured in a good illumination condition. It captures a serious concern when deploying them into real-world low light environments. As known that low light images especially dark images suffer from poor visibility and high noise, and thus only a little or non-useful information can be used  to perform high level perception from them even using powerful deep neural networks.
Therefore, it's necessary to enhance low light images in advance.

Recently, deep learning based methods have been continuously proposed to enhance low light images. These methods learn a convolutional network with paired low light and corresponding normal light images in a supervised fashion. 
Although we have seen great progress made by them, there are mainly three problems that hinder the real-world deployment of those learning based methods. 1) First, it's a challenge to simultaneously acquire low light images from real-world scenes with their corresponding normal light images. Alternatively, researchers introduce to use synthesized low light images, however, the model learned from them cannot be directly deployed into real-world scenarios due to domain shift. 2) Second, it's difficult to deal with extremely low light conditions. Deep learning based methods have demonstrated satisfactory performance for slightly low light images, however, they don't perform well for dark images.
3) Besides, low light images usually suffer from strong noise due to the low signal-to-noise ratio, this also brings a difficulty when enhancing low illumination images.

\begin{figure}[t]
\centering  
\begin{tabular} 
{p{0.135\textwidth}<{\centering}p{0.135\textwidth}<{\centering}p{0.135\textwidth}<{\centering}} 
\IncG[ width=1.06in]{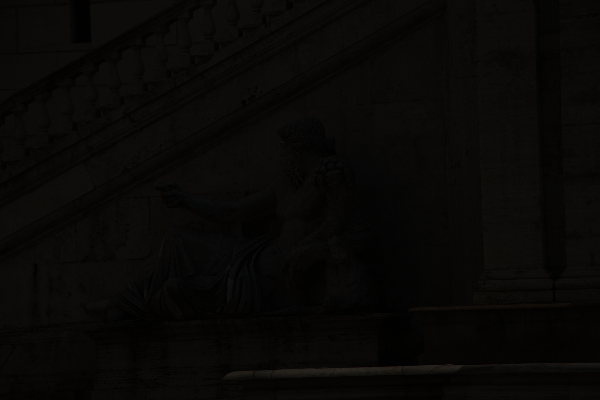}
&\IncG[ width=1.06in]{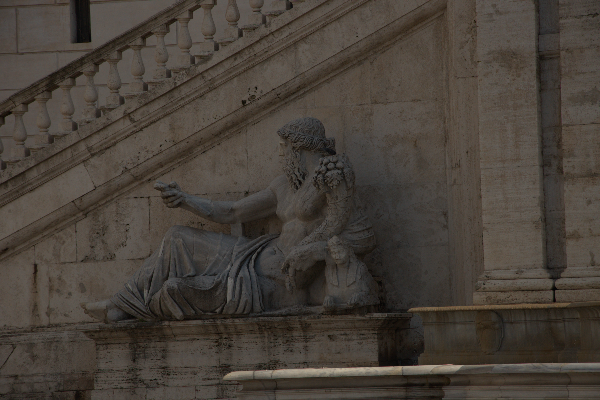}
&\IncG[ width=1.06in]{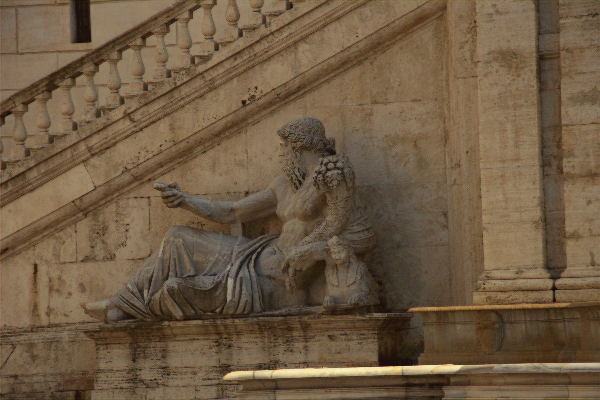}
\\
\IncG[ width=1.06in]{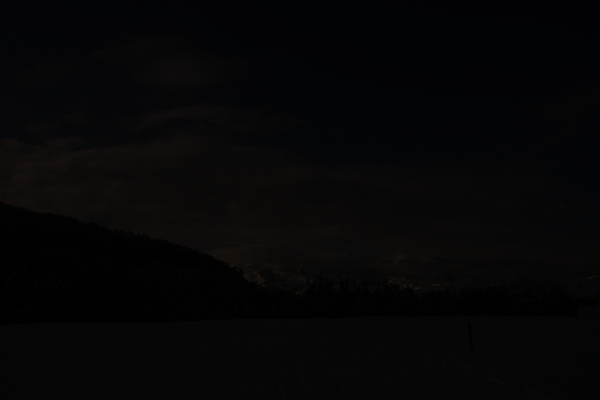}
&\IncG[ width=1.06in]{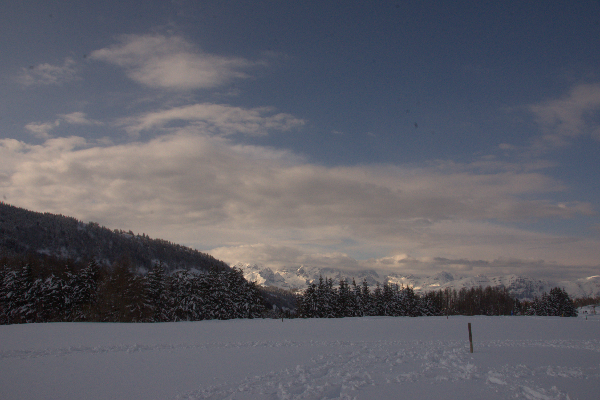}
&\IncG[ width=1.06in]{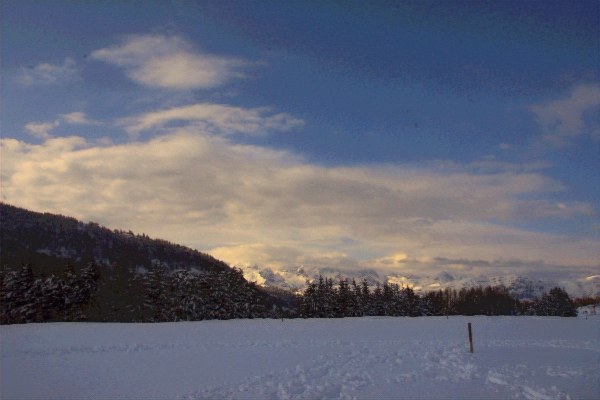}
\\
\IncG[ width=1.06in]{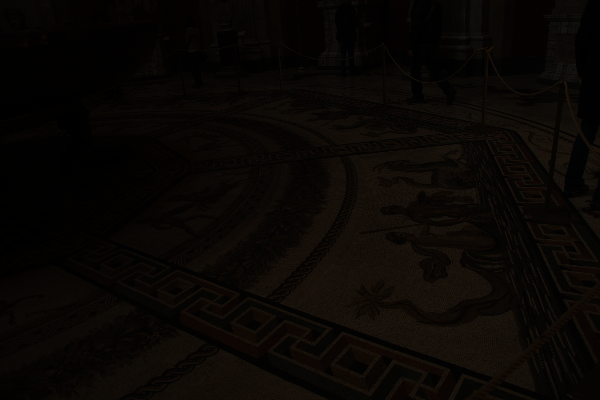}
&\IncG[ width=1.06in]{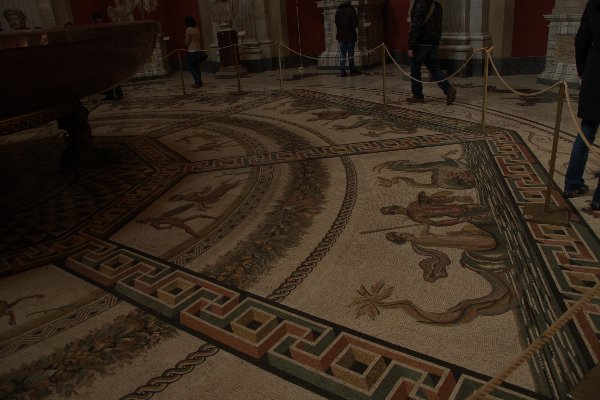}
&\IncG[ width=1.06in]{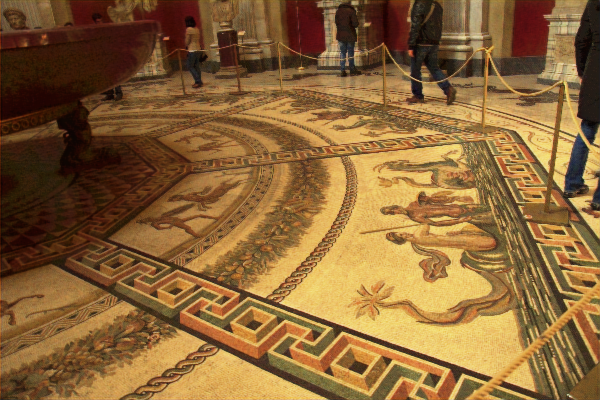}
\\
\footnotesize(a) Low light inputs. &\footnotesize(b) Ground truth.&\footnotesize(c)  Our results.
\end{tabular}
\vspace*{2mm}
\caption{Results of enhancement for three dark images. Our method demonstrates superior performance of enhancement for dark images, as seen that the perceptual quality after enhancement is even better than the ground truth.}
\label{fig_example1}
\end{figure}

\begin{figure*}[!t]
\centering
\subfigure {\includegraphics[scale=0.5]{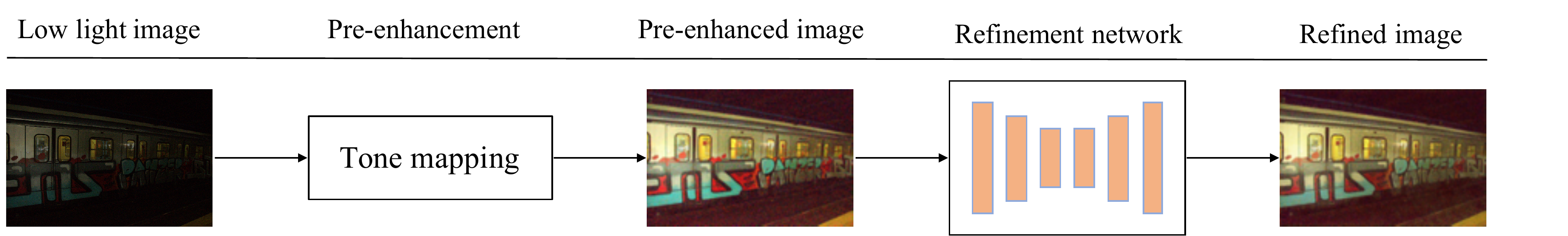}}
\caption{Diagram of the proposed two-stage framework for low light image enhancement. Given a low light image, in the first stage, we employ the tone mapping method proposed in \cite{Ahn2013AdaptiveLT} to pre-enhance the image. In the second stage, we use a refinement network for further improvement of image quality. 
}
\label{fig_arch}
\end{figure*}

Most of the previous studies for low light image enhancement are focused on handling one of the above problems.
For the first problem, researchers have begun to propose unsupervised low light image enhancement approaches. Jiang et al. proposed  EnlightenGAN \cite{jiang2019enlightengan} that enhances low light images with a generative adversarial network. Zhang et al. \cite{zhang2020self} proposed a self-supervised learning based method that can complete the training with even one single low light image based on maximum entropy.
For the second problem, Chen et al.\cite{chen2018learning} propose to recover normal images from extremely dark images by learning a convolutional network  with raw data. 
There are also many approaches have been proposed for denoising of low light images. Remez et al.\cite{remez2017deep} proposed a method that utilizes deep convolutional neural networks for Poisson denoising for low light images. Chatterjee \cite{chatterjee2011noise} et al. used a locally linear embedding framework where a linear embedding is learned for denoising.
It's noted that although previous approaches have demonstrated satisfactory performance for any of the above problems, it would be a difficult challenge when attempting to tackle them at the same time. 
We argue that simultaneously enhancing illumination as well as denoising is a non-trivial problem as they are usually formulated and solved in different paradigms.


To alleviate the above difficulties, in this paper, we decompose the low light into two sub-problems, i.e., the pre-enhancement and post-refinement, and propose a two-stage method to more accurately enhance low light images. 
To be specific, in the first stage, we enhance the illumination map decomposed from a low light image based on the Retinex theory. We employ a tone mapping based method \cite{Ahn2013AdaptiveLT} for the purpose. In the second stage, we design a refinement network to further improve the image quality from the pre-enhanced image obtained in the first stage. We design a comprehensive loss function that combines the loss of image content, perceptual quality, total variation, and adversarial loss.
This stage contributes to the improvement of image quality, especially for noise suppression. Our two-stage strategy demonstrates satisfactory performance even for dark image inputs, an example is given in Fig.\ref{fig_example1} where the results are even better than the ground truth. 

To summarize, the main contribution of this paper is the proposal of a simple two-stage unsupervised approach that performs pre-enhancement and post-refinement for low light image enhancement. It outperforms state-of-the-art methods, including both supervised learning based methods and unsupervised learning based methods on four benchmark datasets. Furthermore, we show two applications of our method in which we
demonstrate that it can archive much accurate feature points matching and can be further seamlessly applied to SLAM in low light conditions. 


\section{Related Works}
\subsection{Traditional Methods}
Traditional approaches can basically be separated into two categories: histogram equalization based methods and Retinex theory based methods. Among them, 
histogram equalization \cite{Coltuc2006ExactHS,elik2011ContextualAV} are the most simply and widely used methods. There are also many Retinex based approaches.  
Guo et al. proposed a method called LIME \cite{Guo2017LIMELI} which first initializes the illumination map with the maximum value in its RGB channels, then imposes a structure prior on the illumination map. 
\cite{Li2018StructureRevealingLI} proposed a robust Retinex model that formulates low light image enhancement as an optimization problem. They additionally applied the $l_1$ norm on the illumination map to constrain the piece-wise smoothness of the illumination. However, these traditional approaches tend to cause color distortion and amply noise in enhanced images.


\subsection{Supervised Based Methods}
\cite{Shen2017MSRnetLI} proposed to learn a deep convolutional network that directly formulates the low light image enhancement as a machine learning problem. The network is learned by penalizing the error between low light images and their corresponding normal light images.
\cite{ren2019low} proposed a two-stream framework which  consists of a content stream network and an edge stream network. \cite{Chen2018Retinex} proposed to use a neural network to decompose a low light image into two components, i.e. an illumination map and a reflectance map based on the Retinex theory, then the enhancement is applied on the two components with ground truth illumination and reflectance map. A similar idea is also adopted in \cite{zhang2019kindling}, where a more accurate network is introduced.
It's noted that supervised learning based methods have brought significant progress on the task, however, the need of image pairs of low light and normal light images for learning makes them hard to be applied to real-world scenarios.

\subsection{Unsupervised Based Methods}
Unsupervised based methods attempt to enhance low light images without pairs of low light and normal light images. To this end,
\cite{lore2017llnet} proposed a deep auto-encoder based approach that learns to enhance from low light images in an unsupervised fashion where the low light images are synthesized with different dark conditions.
Previous methods have also attempted to utilize generative adversarial network (GAN). \cite{jiang2019enlightengan} proposed EnlightenGAN which can be trained in an end-to-end fashion, it achieved competitive performance compared with supervised learning based methods. 
\cite{xiong2020unsupervised} further proposed decoupled networks where illumination enhancement and noise reduction are handled with contrast enhancement and image denoising network, respectively. 
Besides, Zhang et al.\cite{zhang2020self} assumed that the maximum channel of the reflectance should conform to the maximum channel of the low light image and has the maximum entropy. Based on the assumption, they introduced a maximum entropy based Retinex model which can be trained with low light images only. However, the method didn't demonstrate competitive performance against others such as EnlightenGAN.

\section{METHODOLOGY}

As discussed above, it's difficult to get satisfactory performance by directly formulating the low light image enhancement as a learning problem considering the difficulty of simultaneous illumination enhancement and denoising. Therefore, we propose a two-stage framework that performs pre-enhancement and post-refinement to gain better performance. The proposed framework is shown in Fig.~\ref{fig_arch}.
Given a low light image, we first enhance an illumination map decomposed from the low light input. Then the pre-enhanced image is inputted to a refinement network to further suppress noise and improve the overall quality.
The details of our two-stage method are shown below.

\subsection{Pre-enhancement}
According to Retinex theory, an image can be decomposed into an illumination map and a reflectance map, i.e.,
\begin{equation}
    X = I \circ R,
\end{equation}
where $X$ is an RGB image, $I$ and $R$ are illumination and reflectance map, respectively. 
In the first stage, we employ the adaptive tone mapping  \cite{Ahn2013AdaptiveLT} to enhance the illumination map. 
It's represented as:
\begin{equation}
    Y' = \frac{L_g}{L_w} \circ X 
\end{equation}
where  $Y'$ denotes the pre-enhanced image from $X$, $L_w$ is the gray scale of $X$; $L_g$ is the global adaptation output, it is calculated by:
\begin{equation}
    L_g = \frac{\log(L_w/\overline{L}_w + 1)}{\log(L_{wmax}/\overline{L}_w + 1)},
\end{equation}
where $L_{wmax}$ denotes the maximum of $L_w$. $\overline{L}_w$ is  the log-average luminance which can be formulated as:
\begin{equation}
    \overline{L}_w = \exp{(\frac{1}{m * n} \sum(\log(\sigma + L_w)))}
\end{equation}
where $m,n$ denotes the width and height of image, $\sigma$ is a small constant number.

\begin{figure}[t]
\centering  
\begin{tabular} 
{p{0.21\textwidth}<{\centering}p{0.21\textwidth}<{\centering}} 
\IncG[ width=1.6in]{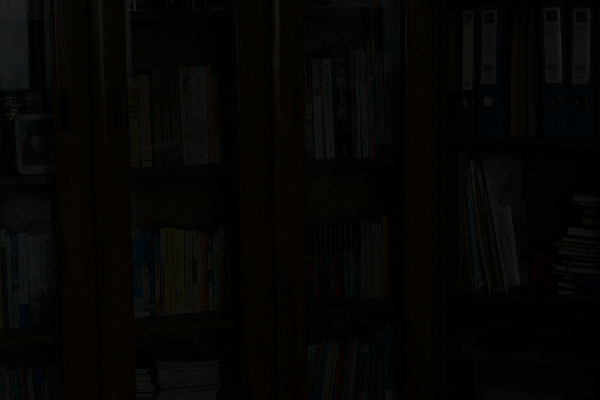}
&\IncG[ width=1.6in]{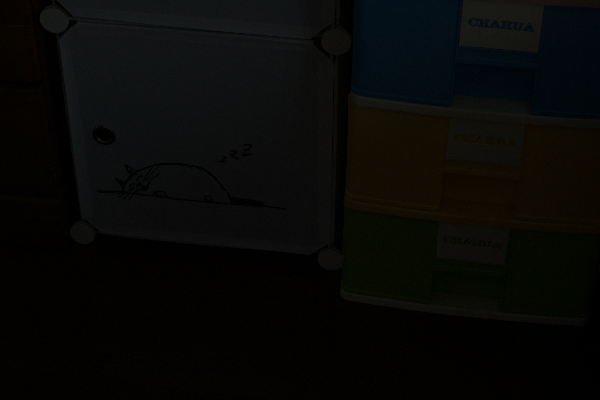}
\\
\IncG[ width=1.6in]{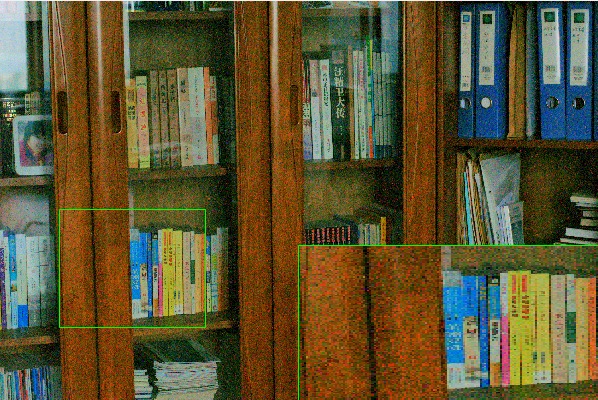}
&\IncG[ width=1.6in]{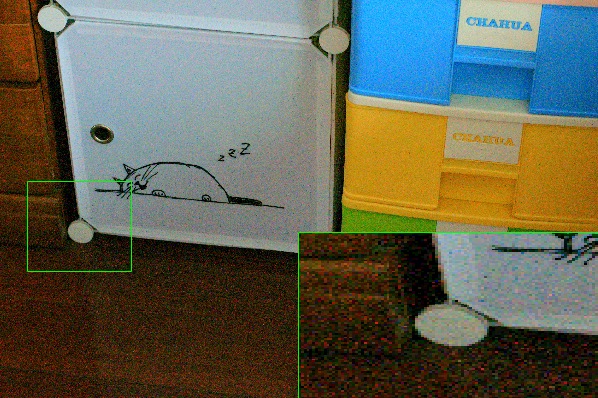}
\\
\IncG[ width=1.6in]{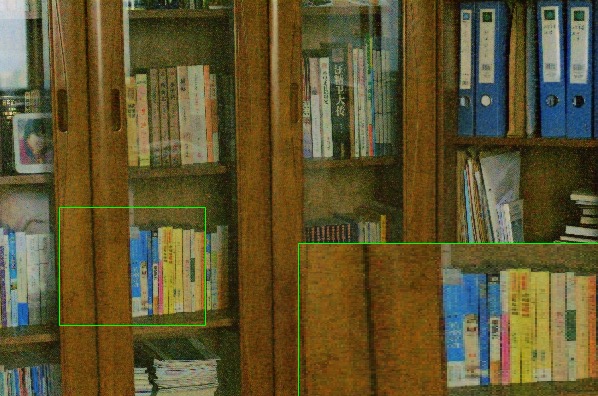}
&\IncG[ width=1.6in]{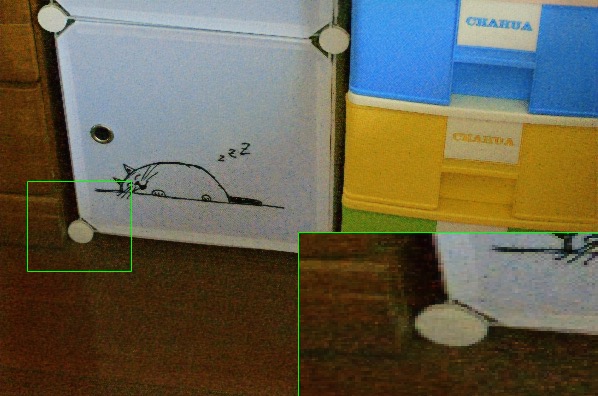}
\\
\end{tabular}
\vspace*{2mm}
\caption{The enhanced results from dark images. The first row denotes the original low light images, the second row shows the results of the pre-enhancement. The third row denotes the results of the two-stage method. It's clear that the post-refinement can effectively suppress noise.}
\label{fig_noise}
\end{figure}

Note that the pre-enhancement can yield competitive performance compared with many deep learning based approaches in terms of illumination enhancement. However, on the other hand, it will largely amplify noise, as seen in the second row of Fig.~\ref{fig_noise}. To cope with this problem, we employ a network for further refinement to improve image quality.


\begin{table}[t]
\centering  
\caption{Input/output, sizes of output features, and input/output channels of each layer for the refinement network on the training set of Unpaired Enhancement Dataset.}
\begin{tabular}{l|p{.08\textwidth}<{\centering}|p{.1\textwidth}<{\centering}|p{.05\textwidth}<{\centering}|p{.05\textwidth}<{\centering}} 
\hline 
Layer &Input/Output  &Output Size   &Input/C &Output/C \\ 
\hline
conv1 & $Y'/x_1$ &128$\times$128 &3 &32 \\
conv2 &$x_1/x_2$ &128$\times$128 &32 &32 \\
down1 &$x_2/x_3$  &64$\times$64  &32 &32 \\  
down2 &$x_3/x_4$ &32$\times$32  &32 &64 \\  
down3 &$x_4/x_5$ &16$\times$16  &64 &128 \\ 
down4 &$x_5/x_6$ &8$\times$8  &128 &256 \\  
conv3 &$x_6/x_7$ &8$\times$8 &256 &512 \\ 
conv4 &$x_7/x_8$ &8$\times$8 &512 &512 \\ 
up1   &$x_8/x_9$ &16$\times$16 &512 &256 \\ 
fusion1 &$x_9, x_5/x_{10}$ &16$\times$16 &384 &256 \\ 
up2  &$x_{10}/x_{11}$ &32$\times$32 &256 &128 \\  
fusion2 &$x_{11}, x_4/x_{12}$  &32$\times$32 &192 &128 \\ 
up3 &$x_{12}/x_{13}$ &64$\times$64 &128 &64 \\ 
fusion3 &$x_{13}, x_3/x_{14}$ &64$\times$64 &96 &64 \\ 
up4 &$x_{14}/x_{15}$ &128$\times$128 &64 &32 \\ 
fusion4 &$x_{15}, x_2/x_{16}$ &128$\times$128 &64 &32 \\ 
conv5 &$x_{16}/Y$ &128$\times$128 &32 &3 \\ 
\hline
\end{tabular}
\label{table_layer}
\end{table}

\subsection{Post-refinement}
The refinement network is an encoder-decoder network which is built on U-net \cite{Ronneberger2015UNetCN}. The encoder consists of four convolutional layers, four downsampling layers. The downsampling layer consists of two convolutional layers followed by a max pooling layer. The encoder extracts features at multiple scales: $1/4$, $1/8$, $1/16$, and $1/32$. 
The decoder employs four upsampling layers to gradually up-scale the final features from the encoder and yields the final output with a convolutional layer. 
For upsampling, we employ the upsampling strategy used in \cite{laina2016deeper, Hu2019RevisitingSI}. 
The details of the refinement net are given in Table~\ref{table_layer}, where conv1 to conv5 are convolutional layers, down1 to down4 are downsampling layers, up1 to up4 are upsampling layers, respectively; Layers of fusion1 to fusion4 are used to concatenate and fuse the features of encoder layers and decoder layers at multi-scales. It consists of two convolutional layers.

As the difficulty to obtain the paired images of low light and normal light in real-world applications, 
we design a comprehensive loss function that can be used to train the network in an unsupervised fashion. 
The loss function consists of four loss terms. The first term is a reconstruction loss that minimizes the pixel-wise loss of image. 
It ensures the consistency of image contents between the refined image and the pre-enhanced image.
It is represented as:
\begin{equation}
    l_{\rm rec} = \|Y - Y'\|_1,
\end{equation}
where $Y'$ denotes a pre-enhanced image, $Y$ is a refined image from $Y'$, it is calculated by the refinement network $N$, i.e. $Y= N(Y')$.
In addition, we employ a perceptual loss to constrain the loss in feature space of VGG \cite{simonyan2014very},
it is represented as:
\begin{equation}
    l_{\rm per} = \|\phi(Y) - \phi(Y')\|_2,
\end{equation}
where $\phi$ denotes VGG network, $\phi(Y')$ is the feature maps extracted from $Y'$.
The reconstruction loss and perceptual loss work in a complementary fashion to avoid color distortion and loss of image contents.

To suppress noise, we additionally apply total variation to the refined image,
\begin{equation}
    l_{\rm tv} = \|\nabla{ Y}\|_1,
\end{equation}
$l_{\rm tv}$ contributes to the reduction of noise, however, it will also lead to the blurred effect on image structure. Therefore, we use an adversarial loss to encourage the refined image to be as close as the clear normal light image. Following \cite{jiang2019enlightengan}, we use the relativistic discriminator structure \cite{jolicoeur2018relativistic} as the discriminative network which is fully convolutional and can handle the input with any size. Then the adversarial loss is given by: 
\begin{equation}
    l_{\rm adv} = ((D(Y) - D(\hat{Y})) -1)^2 + (D(\hat{Y}) - D(Y))^2,
\end{equation}
where $D$ is the discriminator, $\hat{Y}$ denotes normal light images.
As a result, the final loss function for training the refinement network is:
\begin{equation}
    L = l_{\rm rec} + \lambda l_{\rm per} + \mu l_{\rm tv} + \beta l_{\rm adv},
\end{equation}
where $\lambda$, $\mu$ and $\beta$ are weighting coefficients.


\section{EXPERIMENTS}

\begin{figure*}[t]
\centering  
\begin{tabular}
{p{0.05\textwidth}<{\centering}p{0.17\textwidth}<{\centering}p{0.17\textwidth}<{\centering}p{0.17\textwidth}<{\centering}p{0.17\textwidth}<{\centering}} 
(1)
&\IncG[ width=1.3in]{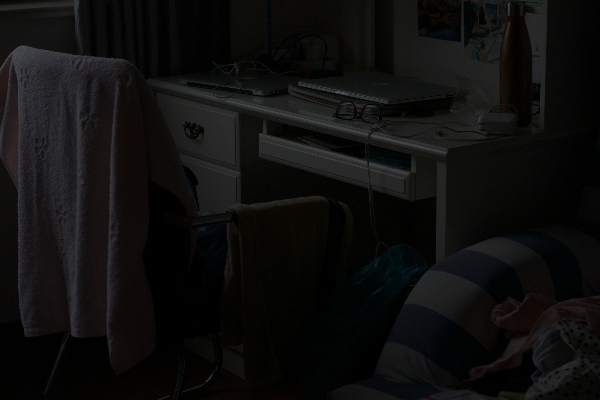}
&\IncG[ width=1.3in]{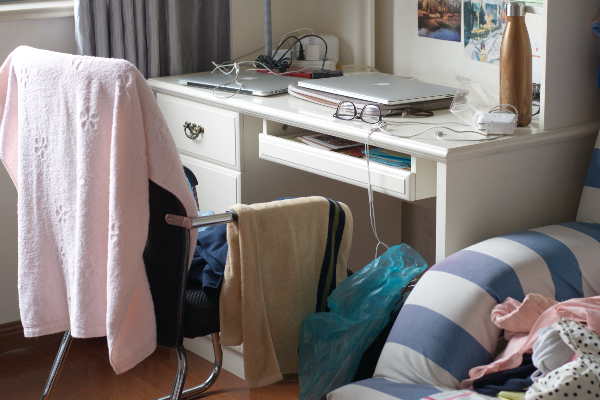}
&\IncG[ width=1.3in]{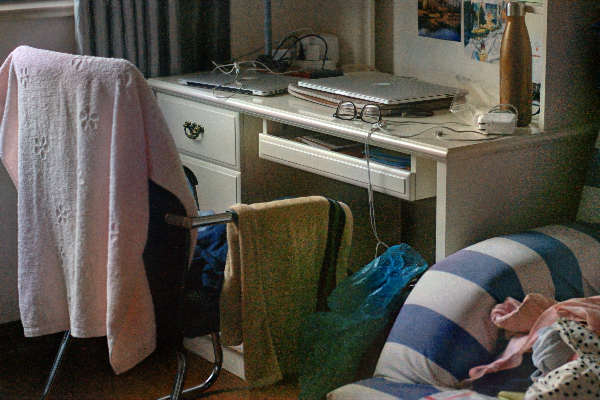}
&\IncG[ width=1.3in]{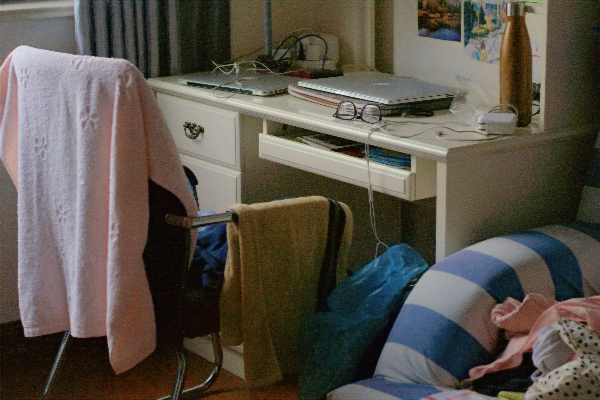}
\\
(2)
&\IncG[ width=1.3in]{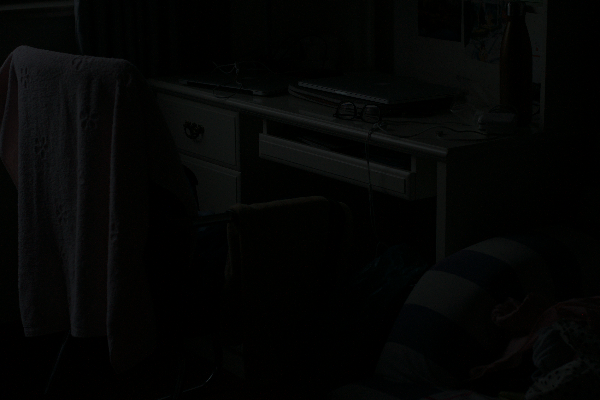}
&\IncG[ width=1.3in]{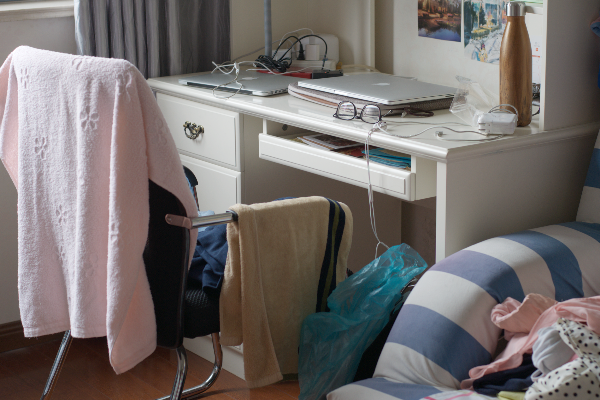}
&\IncG[ width=1.3in]{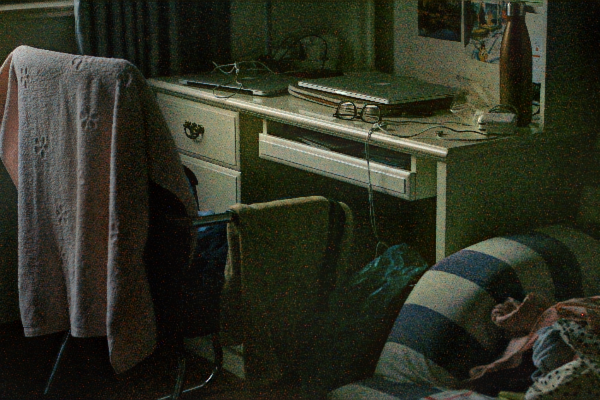}
&\IncG[ width=1.3in]{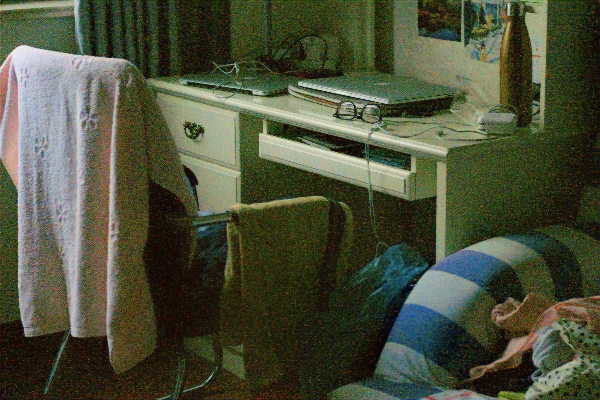}
\\
(3)
&\IncG[ width=1.3in]{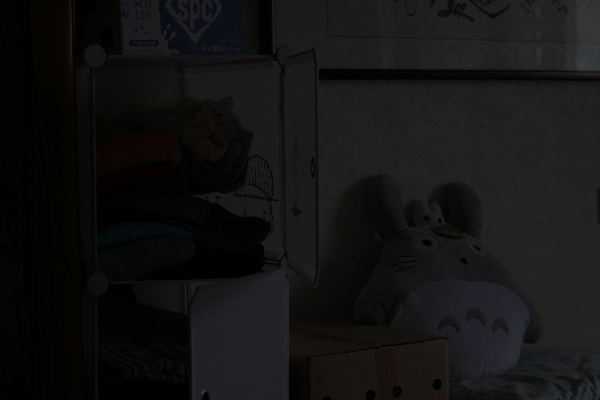}
&\IncG[ width=1.3in]{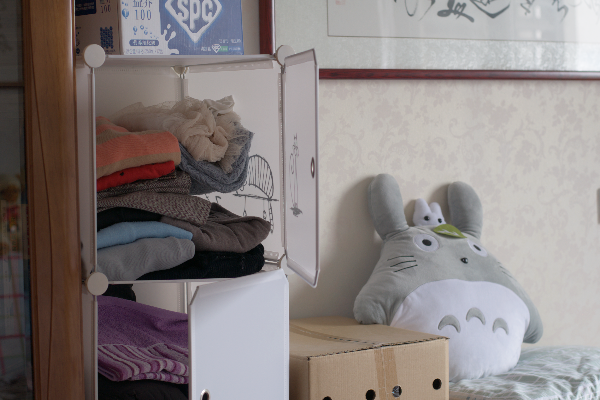}
&\IncG[ width=1.3in]{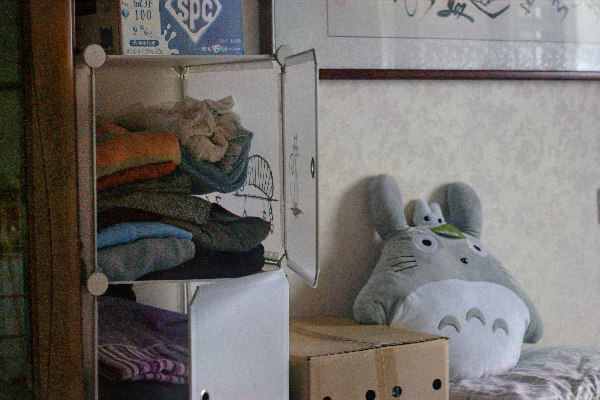}
&\IncG[ width=1.3in]{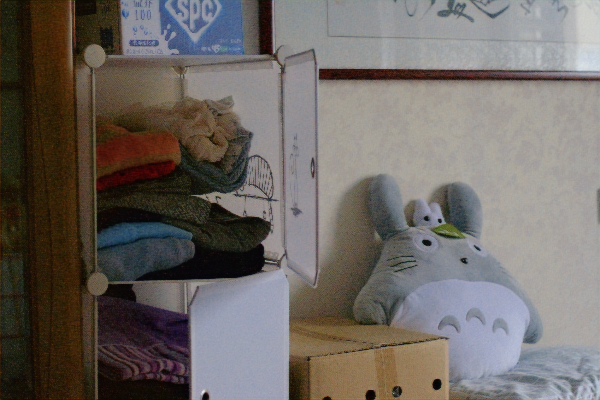}
\\
(4)
&\IncG[ width=1.3in]{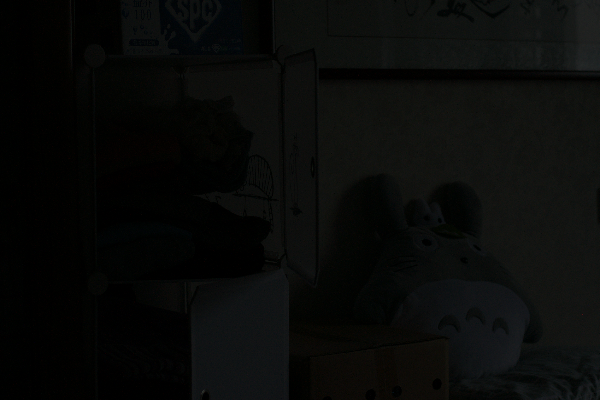}
&\IncG[ width=1.3in]{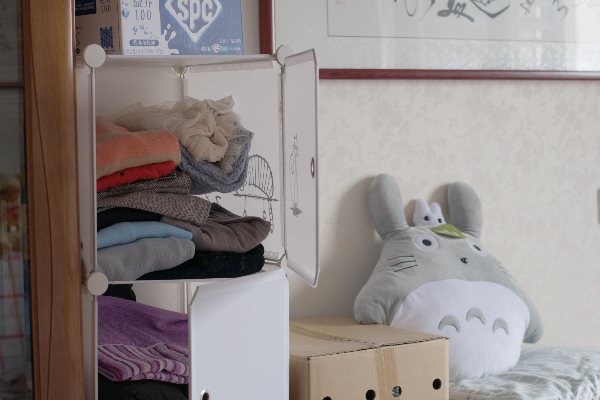}
&\IncG[ width=1.3in]{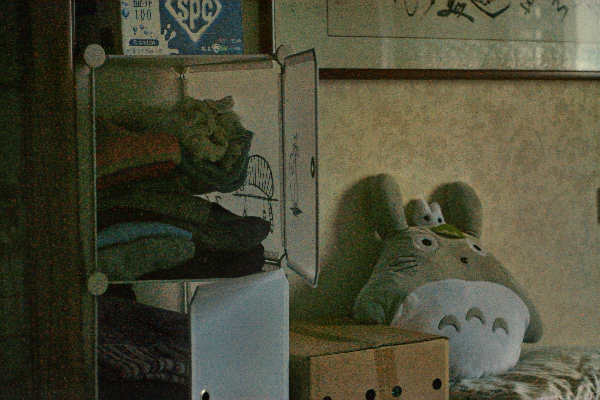}
&\IncG[ width=1.3in]{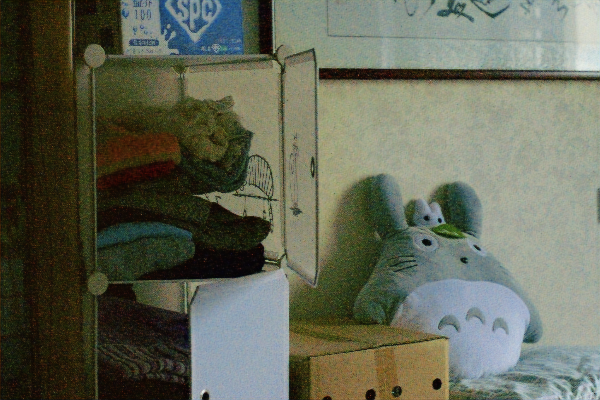}
\\
(5)
&\IncG[ width=1.3in]{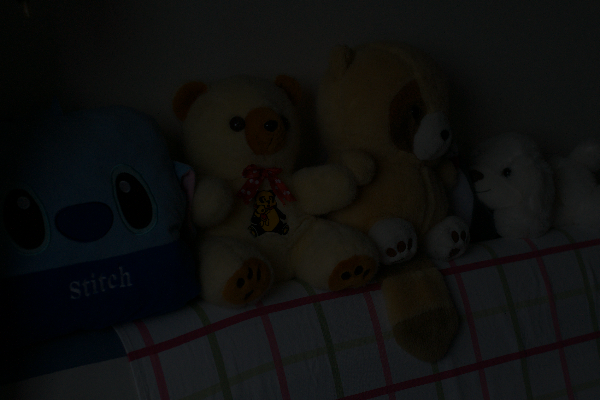}
&\IncG[ width=1.3in]{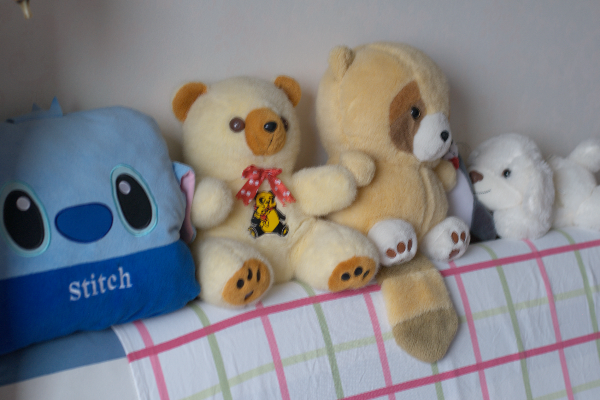}
&\IncG[ width=1.3in]{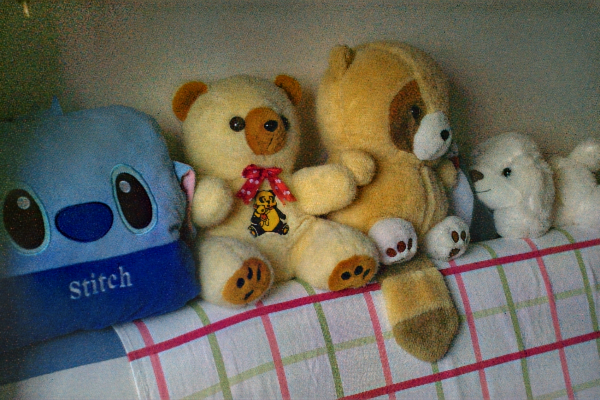}
&\IncG[ width=1.3in]{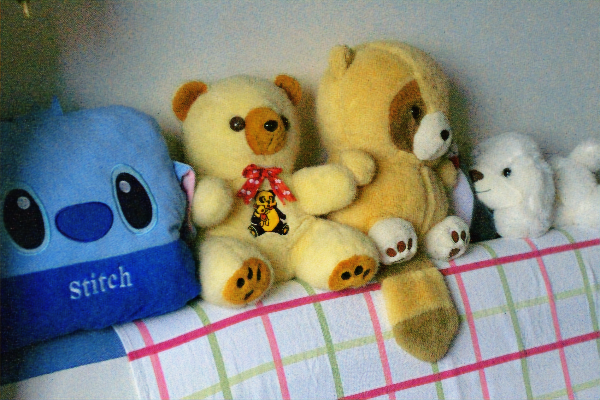}
\\
(6)
&\IncG[ width=1.3in]{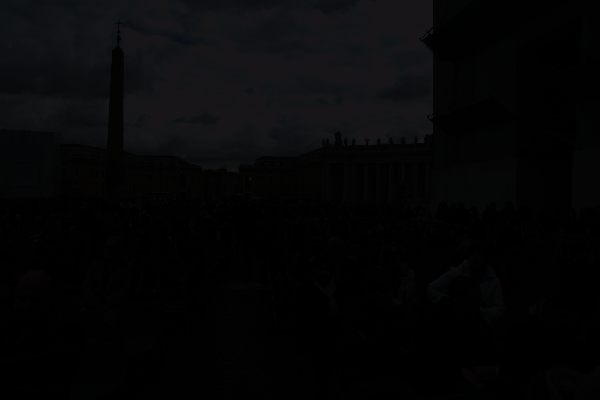}
&\IncG[ width=1.3in]{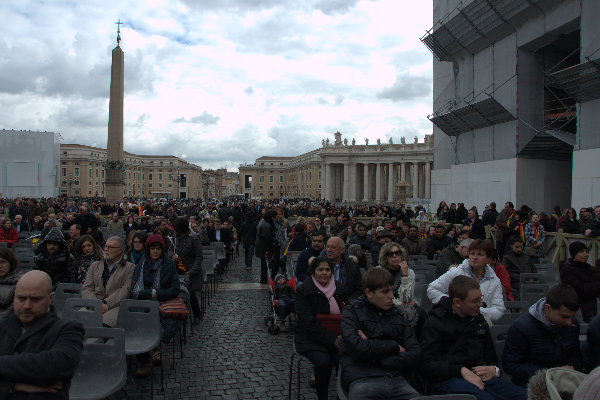}
&\IncG[ width=1.3in]{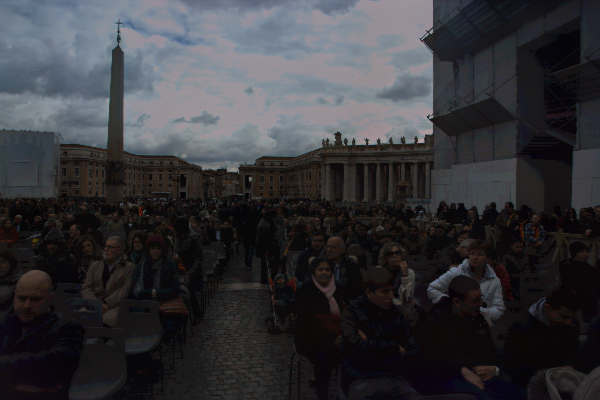}
&\IncG[ width=1.3in]{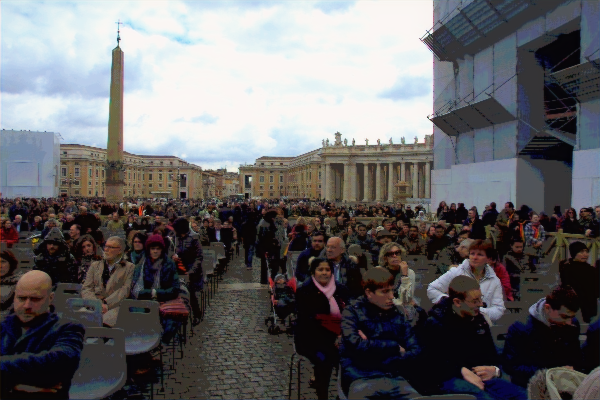}
\\
&(a) Input images & (b) Ground truth & (c) EnlightenGAN & (d) Ours \\
\end{tabular}
\caption{Qualitative comparison between EnlightenGAN and our method on the unpaired enhancement dataset for different methods. From the left to the right; input low light images, ground truth images, results of EnlightenGAN, and results of our method. (1) - (2), (3) - (4) show the results for same scenes under different light conditions. (5) and (6) show results for other dark scenes. Our method outperforms EnlightenGAN for all eight images. }
\label{fig1}
\end{figure*}

\subsection{Datasets}
\paragraph{Unpaired Enhancement Dataset}
The unpaired enhancement dataset \cite{jiang2019enlightengan} is collected from several public datasets.  
The training set is composed of 914 low light images and 1016 normal light images. The test set is composed of 148 pairs of low light and normal light images.  All the images have a resolution of $600 \times 400$. We compare our method with the benchmark method  \cite{jiang2019enlightengan}, i.e. EnlightenGAN on this dataset. 



\paragraph{Benchmark Evaluation Datasets}
For a fair comparison with previous methods, we report more quantitative results on real-world benchmark datasets. We
evaluate our method on MEF\cite{Ma2015PerceptualQA}, LIME\cite{Guo2017LIMELI}, NPE\cite{Wang2013NaturalnessPE}. 
The three datasets are frequently used in previous studies for evaluation, in which MEF, LIME, and NPE have 17, 8, and 10 images, respectively. 

\subsection{Implementation Details}
For learning the refinement network, we employ the training set from the unpaired enhancement dataset. During the training phase, we randomly crop $128 \times 128$ patches from the original $640 \times 400$ resolution images pixels.

We use Adam optimizer with a learning rate of 0.0001. We set $\beta_{1}=0.9$, $\beta_{2}=0.999$, and use weight decay of 0.0001.
The weights $\lambda$ of $l_{\rm tv}$, $\mu$ of $l_{\rm gan}$ and  $\gamma$ of $l_{\rm perceptual}$ are set as $\lambda=0.1$, $\mu=1$  and $\gamma=1$. We train the refinement network for 1000 epochs. 
We conducted all the experiments using PyTorch \cite{paszke2017automatic} with batch size of 64. 


\begin{table}[t]
\begin{center}
\caption{Quantitative comparisons on the  unpaired dataset.}
\label{table_com}\small
\begin{tabular}
{|l|p{.06\textwidth}<{\centering}|p{.06\textwidth}<{\centering}|p{.06\textwidth}<{\centering}|}
\hline
 & PSNR $\uparrow$& SSIM$\uparrow$ &NIQE $\downarrow$\\ \hline
Input &10.370 &0.275 &5.299\\\hline
EnlightenGAN \cite{Chen2018Retinex}  &17.314 &0.711 &4.591 \\\hline
Pre-enhancement \cite{Ahn2013AdaptiveLT}  &17.337 &0.698 &7.012 \\\hline
Post-refinement   &\textbf{18.064} &\textbf{0.720} &\textbf{4.474}\\\hline
\end{tabular}
\end{center}
\end{table}


\begin{table}[t] 
\begin{center}
\caption{Quantitative comparisons of different methods on the benchmark datasets.}
\label{table_com2}\small
\begin{tabular}
{|l|p{.06\textwidth}<{\centering}|p{.06\textwidth}<{\centering}|p{.06\textwidth}<{\centering}|}
\hline
 & MEF & LIME &NPE  \\ \hline
Input &4.265 &4.438 &4.319  \\\hline
RetinexNet \cite{Chen2018Retinex}  &4.149 &4.420 &4.485  \\\hline
LIME \cite{Guo2017LIMELI}   &3.720 &4.155 &4.268  \\\hline
SRIE \cite{Fu2016AWV} &3.475 &3.788 &3.986 \\ \hline
NPE \cite{Wang2013NaturalnessPE} &3.524 &3.905 &3.953\\ \hline
GLAD \cite{Wang2018GLADNetLE} &3.344 &4.128 &3.970 \\ \hline
EnlightenGAN \cite{jiang2019enlightengan}  &3.232 &3.719 &4.113  \\\hline
KinD \cite{zhang2019kindling}  &3.343 &3.724 &3.883  \\\hline
Ours &\textbf{3.027} &\textbf{3.599} &\textbf{3.014}  \\\hline
\end{tabular}
\end{center}
\end{table}


\subsection{Performance Comparison}
We first show the quantitative comparison of our method against EnlightenGAN \cite{jiang2019enlightengan} on the unpaired enhancement dataset. Three metrics are adopted for quantitative comparison, which are PSNR, SSIM, and NIQE. For  PSNR and SSIM, a higher value indicates a better quality, while for NIQE, the lower is better. As seen in Table~\ref{table_com}, the pre-enhancement yields a little better PSNR than EnlightenGAN, but the results are a little worse on SSIM, moreover, it is observed a $33.3\%$ error increase of NIQE. On the other hand, our two-stage method achieves the best performance for all metrics, which indicates the superiority of the configuration of pre-enhancement and post-refinement.
Fig.~\ref{fig1} shows the qualitative comparison against EnlightenGAN. It's observed that both EnlightenGAN and our method can achieve satisfactory performance if there are valuable clues that exist in the inputs as seen in Fig.~\ref{fig1} (1) and (3), however, it's difficult to get the same results if the inputs are extremely dark, as seen in Fig.~\ref{fig1} (2) and (4). Nevertheless, our method demonstrates better performance for dark images, as also seen in Fig.~\ref{fig1} (5) and (6).

For more comparisons against other methods, we provide the results of quantitative comparisons on the MEF, LIME, and NPE datasets. It is noted that we do not train a new model for these three datasets. To evaluate the generability of our method, we use the trained model on the unpaired dataset and test it on the MEF, LIME, and NPE datasets. As there are no reference images are available for these datasets, we use the NIQE value as image quality evaluation in compliance with previous methods \cite{jiang2019enlightengan,zhang2019kindling}. We compare our method against RetinexNet \cite{Chen2018Retinex}, LIME \cite{Guo2017LIMELI},SRIE \cite{Fu2016AWV}, NPE \cite{Wang2013NaturalnessPE}, GLAD \cite{Wang2018GLADNetLE}, KinD \cite{zhang2019kindling}, and EnlightenGAN \cite{jiang2019enlightengan}. The numerical results are shown in Table~\ref{table_com2}, it is seen that our method shows a clear advantage against the others as it outperforms them for all datasets.

\begin{figure*}[!t]
\centering
\subfigure {\includegraphics[scale=0.43]{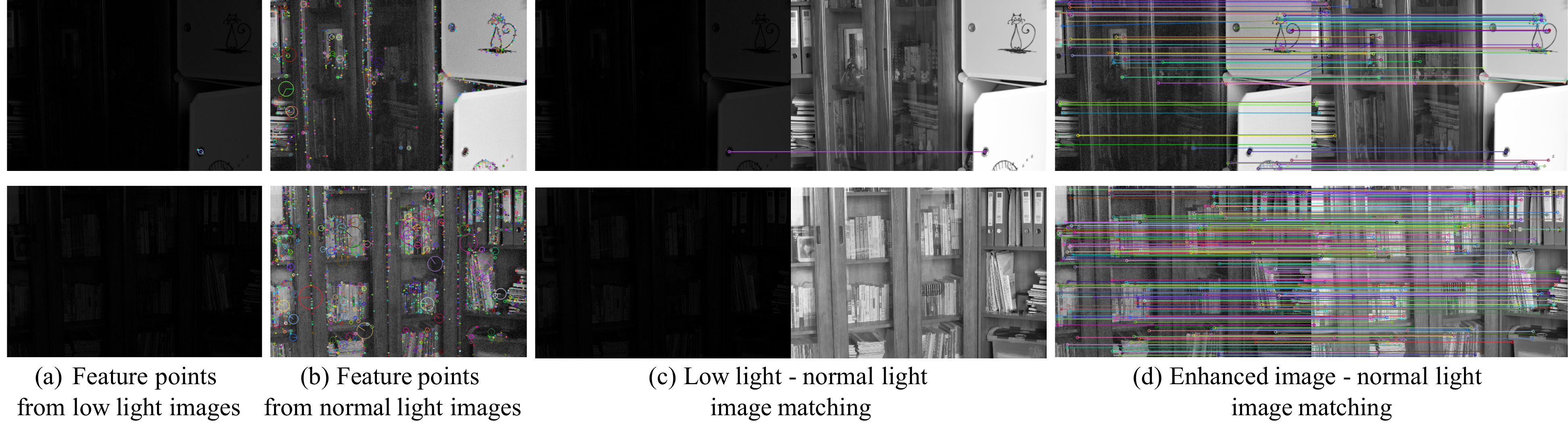}}
\vspace{-3mm}
\caption{Qualitative results of feature points detection and matching. (a) shows results of detected points with SIFT from low light images, (b) shows results of detected points with SIFT from enhanced images of low light images, (c) is the results of image matching between low light images and normal light images, (d) is the results between enhanced images from low light images and normal light images.}
\label{fig_matching}
\end{figure*}

\subsection{Application: Low Light and Normal Light Image Matching}
\label{app_matching}
Image matching is one of the fundamental techniques in robot vision and it plays an indispensable role in many applications such as image retrieval, structure from motion, image based localization, etc. Unsurprisingly, the low light condition easily leads algorithms of feature points matching to malfunction. Zhou et al.\cite{zhou2016evaluating} also discussed the necessity of image matching between a low light image and a normal light image.

We show that our method can be applied to low light and normal light image matching. To be specific, we conduct the image matching between a low light image and its corresponded normal light image on the test set of the unpaired dataset. 
We use SIFT to detect feature points and generate descriptors. Then, they are matched with the 2-nearest neighbor algorithm. To get more accurate matching, we use a small number for distance ratio. In our experiment, we set it to 0.3.
To further eliminate mismatches, we apply the RANSAC algorithm \cite{fischler1981random} to remove outliers.

\begin{table}[t] 
\begin{center}
\caption{Results for feature points detection and matching with and without the enhancement of low light images.}
\label{table_img_matching}\small
\begin{tabular}
{|l|c|c|c|c|}
\hline
&Detected points  &Matches &Match rate \\\hline
Low light images  &22195  &17424 &13.85\% \\ \hline
EnlightenGAN &185930 &38152 &30.32\% \\ \hline
Ours   &172554 &40676 &32.33\% \\ \hline
\end{tabular}
\end{center}
\end{table}

 The quantitative results are given in Table~\ref{table_img_matching}.
As a result, 22195 points are detected from the low light images (there are 125825 feature points detected from the normal light images). From that, we can only get 17424 matches, i.e. the match rate\footnote{the match rate is the rate of the number of final matches divided by points detected from normal light images.} is only $13.85\%$. On the other hand, after applying the enhancement with our method, the number of matched points is 40676 and the match rate is significantly improved from $13.85\%$ to $32.33\%$. It's noted that EnlightenGAN detected more feature points than our method though, the match rate is lower than ours.
It suggests that there are many noisy points detected by EnlightenGAN and the quality of enhanced images by EnlightenGAN are not as good as ours.
A qualitative comparison is shown in Fig.~\ref{fig_matching} (c), as seen that there is almost no successful matching for the original low light images. This is because it's extremely difficult to detect feature points from low light images (Fig.~\ref{fig_matching} (a)).
After applying low light image enhancement (Fig.~\ref{fig_matching} (b)), a large amount of feature points are detected and they can be correctly matched (Fig.~\ref{fig_matching} (d)).


\subsection{Application: SLAM in Low Light Conditions}


\begin{figure*}[!t]
\centering  
\begin{tabular}
{p{.2\textwidth}p{.2\textwidth}p{.2\textwidth}p{.2\textwidth}p{.07\textwidth}l}
\IncG[ width=1.5in]{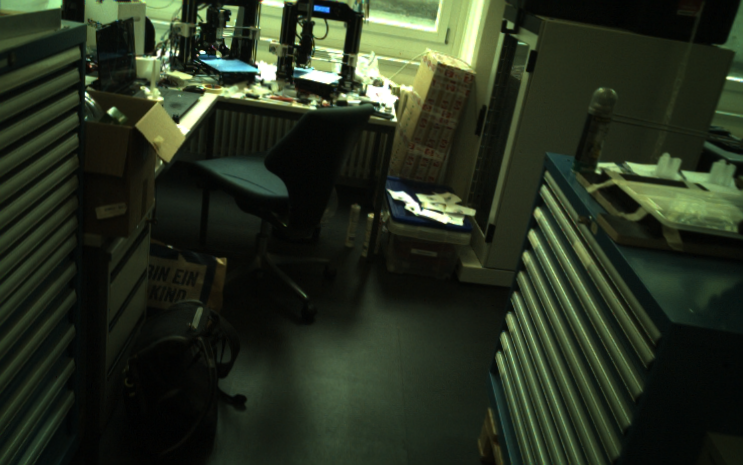}
&\IncG[ width=1.5in]{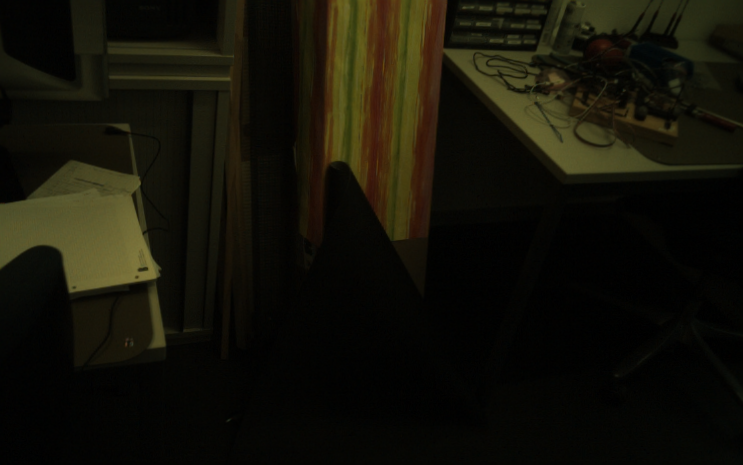}
&\IncG[ width=1.5in]{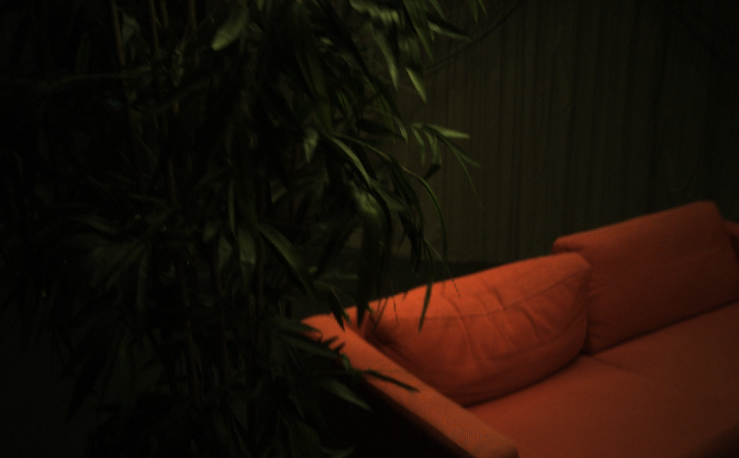}
&\IncG[ width=1.5in]{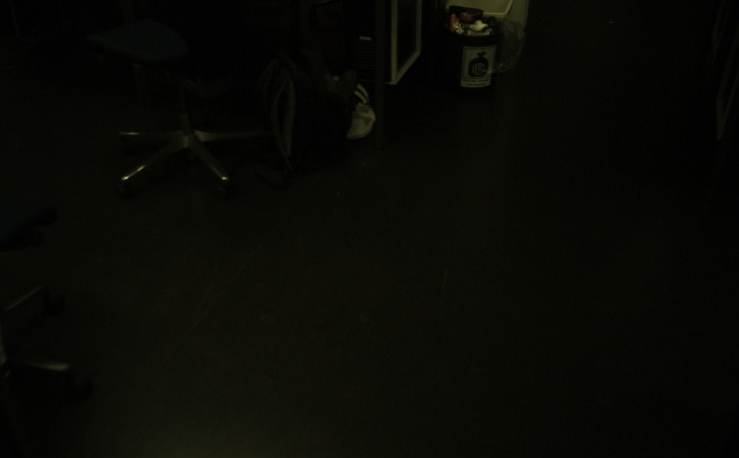}
&\small Original images\\
\IncG[ width=1.5in]{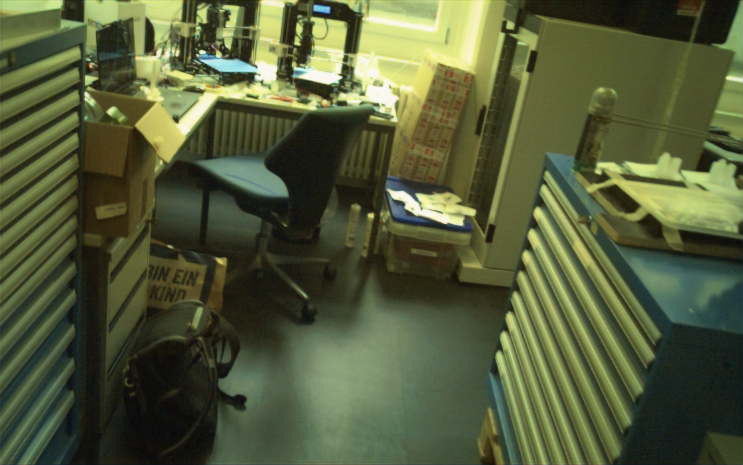}
&\IncG[ width=1.5in]{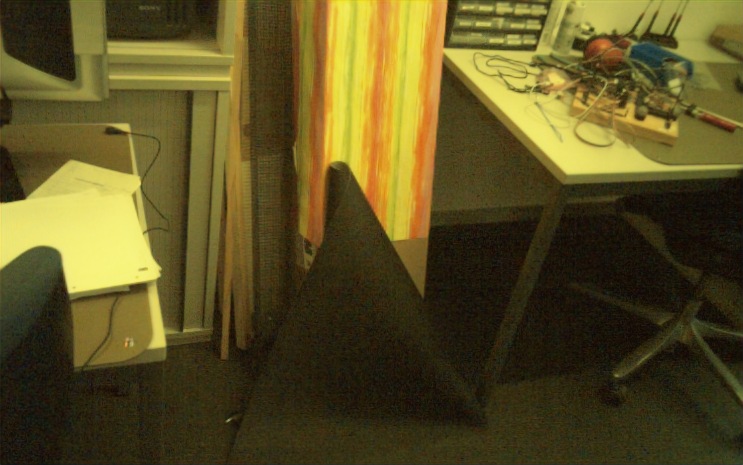}
&\IncG[ width=1.5in]{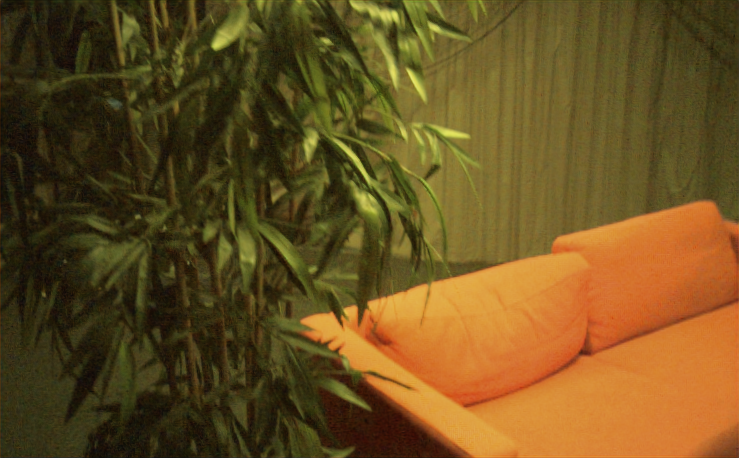}
&\IncG[ width=1.5in]{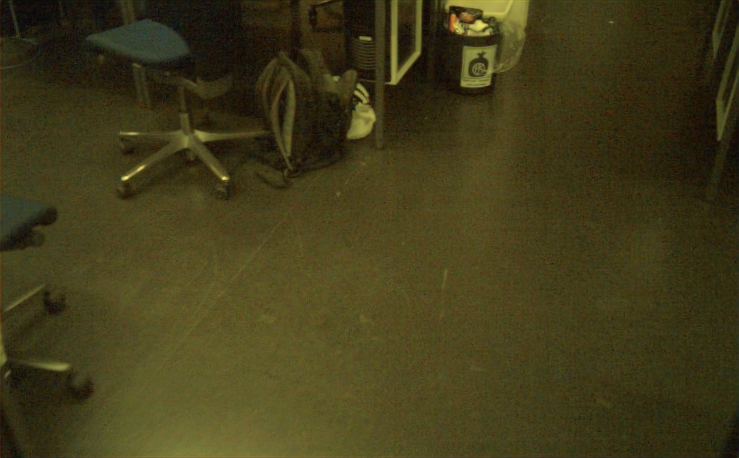}
&\small Enhanced  images\\
\centering
$sfm\_lab\_room\_1$ & \centering $sfm\_lab\_room\_2$ & \centering $plant\_scene\_1$ &\centering $large\_loop\_1$ &
\end{tabular}
\caption{Selected images from ETH3D SLAM benchmark dataset. The first row shows the original low light images, the second row shows the enhanced images with our method.}
\label{fig_eth_img}
\end{figure*}

\begin{figure*}[!t]
\centering
\subfigure {\includegraphics[width=6.2in]{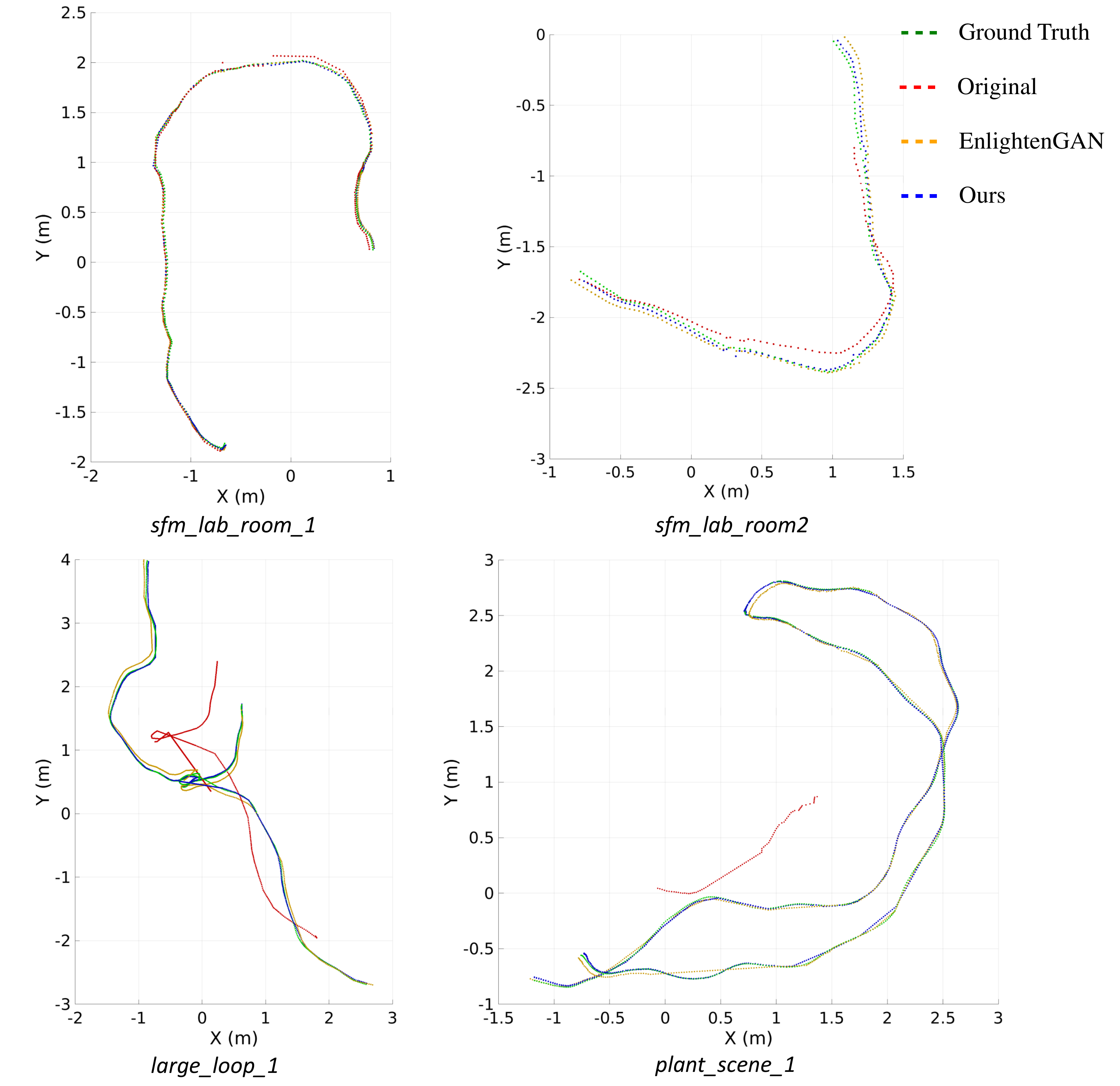}}
\caption{Camera trajectories for the four sequences of ETH3D dataset, where the ground truths are shown in green and the results of our method are shown in blue.
}
\label{fig_res_slam}
\end{figure*}


Vision based monocular SLAM  tends to fail in low light environments.  
To evaluate the application of our method for SLAM, we test it on the ETH3D SLAM benchmark \cite{Schps2019BADSB}. Specifically, we use the ORB-SLAM2 \cite{mur2017orb} to perform RGBD based monocular SLAM. We evaluate our method as well as EnlightenGAN  $sfm\_lab\_room\_1$, $sfm\_lab\_room\_2$, $large\_loop\_1$ and $plant\_scene\_1$ taken from ETH3D SLAM benchmark \cite{Schps2019BADSB}. They are captured in low light conditions but not completely dark. An example is given in Fig.~\ref{fig_eth_img}, where the first row shows the original low light images taken from the above sequences, and the second row shows the images enhanced by our method.

\begin{table}[!t] 
\begin{center}
\caption{$SE3\_ATE\_RMSE (cm)$ on $sfm\_lab\_room\_1$, $sfm\_lab\_room\_2$, $large\_loop\_1$ and $plant\_scene\_1$.}
\label{table_slam}\small
\begin{tabular}
{|l|p{.07\textwidth}<{\centering}|p{.11\textwidth}<{\centering}|p{.07\textwidth}<{\centering}|}
\hline
&Original &EnlightenGAN &Ours \\ \hline
$sfm\_lab\_room\_1$ &3.134&1.907 &1.764 \\ \hline
$sfm\_lab\_room\_2$ &Fail &5.824 &2.956 \\ \hline
$large\_loop\_1$  &Fail &10.401 &4.552 \\ \hline
$plant\_scene\_1$ &Fail &3.356 &1.428 \\ \hline
\end{tabular}
\end{center}
\end{table}


 Table~\ref{table_slam} shows the quantitative comparisons.  It's seen that ORB-SLAM2 only successfully performed on $sfm\_lab\_room\_1$\footnote{ The $SE3\_ATE\_RMSE$ is 1.850 given in the official benchmark while the result is 3.134 performed by us. The reason is considered as the difference of setting of parameters for ORB-SLAM2.} without the low light image enhancement. However it  failed on $sfm\_lab\_room\_2$, $large\_loop\_1$ and $plant\_scene\_1$ which is consistent with the results given in the benchmark \cite{Schps2019BADSB}. 
 On the other hand, the SLAM can be improved significantly if we apply low light image enhancement. As seen that
 our method performs better than EnlightenGAN for all of these four sequences. It is slightly better on $sfm\_lab\_room\_1$ and outperforms EnlightenGAN  on $sfm\_lab\_room\_2$, $large\_loop\_1$ and $plant\_scene\_1$ by a good margin (achieving $49.50\%$, $56.23\%$ and $57.45\%$ improvement of the accuracy). 
 In our experiments, our method takes 95 ms to enhance a $739 \times 458$ resolution image, it can ensure SLAM to be performed in real-time. Fig.~\ref{fig_res_slam} shows camera trajectories for different inputs. As there are several pieces of ground truth trajectories are missing for the sequences of $plant\_scene\_1$ and $large\_loop\_1$, we only show the correct ground truths. They are shown in green and the results from the original images, enhanced images by EnlightenGAN and our method are shown in red, orange and blue, respectively.


\section{Conclusion}
In this paper,  we revisited the problem of real-world low light image enhancement. We point out that there are mainly three challenges that hinder the deployment of most of previous learning based methods. The first is the need of low light and normal light image pairs for learning. To overcome this difficulty, we proposed an unsupervised method that can be implemented with unpaired images using adversarial training. Other challenges are the difficulty of handling very dark input images and the poor ability of denoising. To alleviate the difficulties, we take a two-stage strategy that first pre-enhances a low light image and further refines it with a refinement network. 
Experimental results show that our two-stage approach outperforms previous methods on four benchmark datasets.
We argue that the proposed method can be used as an effective image pre-processing tool for low light image enhancement. In experiments, we demonstrate two useful applications of our method. The first is image matching and the second is SLAM. We show that both of them are vulnerable to low light conditions, nevertheless, they can be significantly improved with the image enhancement performed by our method. In the future, we will speed up our method with some compression techniques for deep neural networks and explore more applications for the perception of robots.
\bibliographystyle{IEEEtranS}
\bibliography{egbib}

\begin{thebibliography}{10}
\providecommand{\url}[1]{#1}
\csname url@samestyle\endcsname
\providecommand{\newblock}{\relax}
\providecommand{\bibinfo}[2]{#2}
\providecommand{\BIBentrySTDinterwordspacing}{\spaceskip=0pt\relax}
\providecommand{\BIBentryALTinterwordstretchfactor}{4}
\providecommand{\BIBentryALTinterwordspacing}{\spaceskip=\fontdimen2\font plus
\BIBentryALTinterwordstretchfactor\fontdimen3\font minus
  \fontdimen4\font\relax}
\providecommand{\BIBforeignlanguage}[2]{{%
\expandafter\ifx\csname l@#1\endcsname\relax
\typeout{** WARNING: IEEEtranS.bst: No hyphenation pattern has been}%
\typeout{** loaded for the language `#1'. Using the pattern for}%
\typeout{** the default language instead.}%
\else
\language=\csname l@#1\endcsname
\fi
#2}}
\providecommand{\BIBdecl}{\relax}
\BIBdecl

\bibitem{Ahn2013AdaptiveLT}
H.~Ahn, B.~Keum, D.~Kim, and H.~S. Lee, ``Adaptive local tone mapping based on
  retinex for high dynamic range images,'' \emph{2013 IEEE International
  Conference on Consumer Electronics (ICCE)}, pp. 153--156, 2013.

\bibitem{8025618}
G.~{Bresson}, Z.~{Alsayed}, L.~{Yu}, and S.~{Glaser}, ``Simultaneous
  localization and mapping: A survey of current trends in autonomous driving,''
  \emph{IEEE Transactions on Intelligent Vehicles}, vol.~2, no.~3, pp.
  194--220, 2017.

\bibitem{chatterjee2011noise}
P.~Chatterjee, N.~Joshi, S.~B. Kang, and Y.~Matsushita, ``Noise suppression in
  low-light images through joint denoising and demosaicing,'' in
  \emph{Proceedings of the IEEE Conference on Computer Vision and Pattern
  Recognition}.\hskip 1em plus 0.5em minus 0.4em\relax IEEE, 2011, pp.
  321--328.

\bibitem{chen2018learning}
C.~Chen, Q.~Chen, J.~Xu, and V.~Koltun, ``Learning to see in the dark,'' in
  \emph{Proceedings of the IEEE Conference on Computer Vision and Pattern
  Recognition}, 2018, pp. 3291--3300.

\bibitem{Chen2018Retinex}
W.~Y. Chen~Wei, Wenjing~Wang and J.~Liu, ``Deep retinex decomposition for
  low-light enhancement,'' in \emph{British Machine Vision Conference}.\hskip
  1em plus 0.5em minus 0.4em\relax British Machine Vision Association, 2018.

\bibitem{Coltuc2006ExactHS}
D.~Coltuc, P.~Bolon, and J.-M. Chassery, ``Exact histogram specification,''
  \emph{IEEE Transactions on Image Processing}, vol.~15, pp. 1143--1152, 2006.

\bibitem{fischler1981random}
M.~A. Fischler and R.~C. Bolles, ``Random sample consensus: a paradigm for
  model fitting with applications to image analysis and automated
  cartography,'' \emph{Communications of the ACM}, vol.~24, no.~6, pp.
  381--395, 1981.

\bibitem{Fu2016AWV}
X.~Fu, D.~Zeng, Y.~Huang, X.~Zhang, and X.~Ding, ``A weighted variational model
  for simultaneous reflectance and illumination estimation,'' \emph{2016 IEEE
  Conference on Computer Vision and Pattern Recognition (CVPR)}, pp.
  2782--2790, 2016.

\bibitem{Guo2017LIMELI}
X.~Guo, Y.~Li, and H.~Ling, ``Lime: Low-light image enhancement via
  illumination map estimation,'' \emph{IEEE Transactions on Image Processing},
  vol.~26, pp. 982--993, 2017.

\bibitem{he2016deep}
K.~He, X.~Zhang, S.~Ren, and J.~Sun, ``Deep residual learning for image
  recognition,'' in \emph{Proceedings of the IEEE conference on computer vision
  and pattern recognition}, 2016, pp. 770--778.

\bibitem{Hu2019RevisitingSI}
J.~Hu, M.~Ozay, Y.~Zhang, and T.~Okatani, ``Revisiting single image depth
  estimation: Toward higher resolution maps with accurate object boundaries,''
  2019.

\bibitem{hu2019visualization}
hunjie Hu, Y.~Zhang, and T.~Okatani, ``Visualization of convolutional neural
  networks for monocular depth estimation,'' in \emph{Proceedings of the IEEE
  International Conference on Computer Vision}, 2019, pp. 3869--3878.

\bibitem{laina2016deeper}
L.~Iro, R.~Christian, B.~Vasileios, T.~Federico, and N.~Nassir, ``Deeper depth
  prediction with fully convolutional residual networks,'' in \emph{3DV}, 2016,
  pp. 239--248.

\bibitem{jiang2019enlightengan}
Y.~Jiang, X.~Gong, D.~Liu, Y.~Cheng, C.~Fang, X.~Shen, J.~Yang, P.~Zhou, and
  Z.~Wang, ``Enlightengan: Deep light enhancement without paired supervision,''
  \emph{arXiv preprint arXiv:1906.06972}, 2019.

\bibitem{jolicoeur2018relativistic}
A.~Jolicoeur-Martineau, ``The relativistic discriminator: a key element missing
  from standard gan,'' \emph{arXiv preprint arXiv:1807.00734}, 2018.

\bibitem{Li2018StructureRevealingLI}
M.~Li, J.~Liu, W.~Yang, X.~Sun, and Z.~Guo, ``Structure-revealing low-light
  image enhancement via robust retinex model,'' \emph{IEEE Transactions on
  Image Processing}, vol.~27, pp. 2828--2841, 2018.

\bibitem{lore2017llnet}
K.~G. Lore, A.~Akintayo, and S.~Sarkar, ``Llnet: A deep autoencoder approach to
  natural low-light image enhancement,'' \emph{Pattern Recognition}, vol.~61,
  pp. 650--662, 2017.

\bibitem{ma2017sparse}
F.~Ma and S.~Karaman, ``Sparse-to-dense: Depth prediction from sparse depth
  samples and a single image,'' \emph{ICRA}, 2018.

\bibitem{Ma2015PerceptualQA}
K.~Ma, K.~Zeng, and Z.~Wang, ``Perceptual quality assessment for multi-exposure
  image fusion,'' \emph{IEEE Transactions on Image Processing}, vol.~24, pp.
  3345--3356, 2015.

\bibitem{milan2018semantic}
A.~Milan, T.~Pham, K.~Vijay, D.~Morrison, A.~W. Tow, L.~Liu, J.~Erskine,
  R.~Grinover, A.~Gurman, T.~Hunn \emph{et~al.}, ``Semantic segmentation from
  limited training data,'' in \emph{2018 IEEE International Conference on
  Robotics and Automation (ICRA)}.\hskip 1em plus 0.5em minus 0.4em\relax IEEE,
  2018, pp. 1908--1915.

\bibitem{milioto2018real}
A.~Milioto, P.~Lottes, and C.~Stachniss, ``Real-time semantic segmentation of
  crop and weed for precision agriculture robots leveraging background
  knowledge in cnns,'' in \emph{2018 IEEE international conference on robotics
  and automation (ICRA)}.\hskip 1em plus 0.5em minus 0.4em\relax IEEE, 2018,
  pp. 2229--2235.

\bibitem{mur2017orb}
R.~Mur-Artal and J.~D. Tard{\'o}s, ``Orb-slam2: An open-source slam system for
  monocular, stereo, and rgb-d cameras,'' \emph{IEEE Transactions on Robotics},
  vol.~33, no.~5, pp. 1255--1262, 2017.

\bibitem{nekrasov2019real}
V.~Nekrasov, T.~Dharmasiri, A.~Spek, T.~Drummond, C.~Shen, and I.~Reid,
  ``Real-time joint semantic segmentation and depth estimation using asymmetric
  annotations,'' in \emph{2019 International Conference on Robotics and
  Automation (ICRA)}.\hskip 1em plus 0.5em minus 0.4em\relax IEEE, 2019, pp.
  7101--7107.

\bibitem{paszke2017automatic}
A.~Paszke, S.~Gross, and A.~Lerer, ``Automatic differentiation in pytorch,''
  2017.

\bibitem{remez2017deep}
T.~Remez, O.~Litany, R.~Giryes, and A.~M. Bronstein, ``Deep convolutional
  denoising of low-light images,'' \emph{arXiv preprint arXiv:1701.01687},
  2017.

\bibitem{ren2019low}
W.~Ren, S.~Liu, L.~Ma, Q.~Xu, X.~Xu, X.~Cao, J.~Du, and M.-H. Yang, ``Low-light
  image enhancement via a deep hybrid network,'' \emph{IEEE Transactions on
  Image Processing}, vol.~28, no.~9, pp. 4364--4375, 2019.

\bibitem{Ronneberger2015UNetCN}
O.~Ronneberger, P.~Fischer, and T.~Brox, ``U-net: Convolutional networks for
  biomedical image segmentation,'' in \emph{MICCAI}, 2015.

\bibitem{Schps2019BADSB}
T.~Sch{\"o}ps, T.~Sattler, and M.~Pollefeys, ``Bad slam: Bundle adjusted direct
  rgb-d slam,'' \emph{2019 IEEE/CVF Conference on Computer Vision and Pattern
  Recognition (CVPR)}, pp. 134--144, 2019.

\bibitem{Shen2017MSRnetLI}
L.~Shen, Z.~Yue, F.~Feng, Q.~Chen, S.~Liu, and J.~Ma, ``Msr-net: Low-light
  image enhancement using deep convolutional network,'' \emph{ArXiv}, vol.
  abs/1711.02488, 2017.

\bibitem{simonyan2014very}
K.~Simonyan and A.~Zisserman, ``Very deep convolutional networks for
  large-scale image recognition,'' \emph{arXiv preprint arXiv:1409.1556}, 2014.

\bibitem{Wang2013NaturalnessPE}
S.~Wang, J.~Zheng, H.~Hu, and B.~Li, ``Naturalness preserved enhancement
  algorithm for non-uniform illumination images,'' \emph{IEEE Transactions on
  Image Processing}, vol.~22, pp. 3538--3548, 2013.

\bibitem{Wang2018GLADNetLE}
W.~Wang, C.~Wei, W.~Yang, and J.~Liu, ``Gladnet: Low-light enhancement network
  with global awareness,'' \emph{2018 13th IEEE International Conference on
  Automatic Face and Gesture Recognition (FG 2018)}, pp. 751--755, 2018.

\bibitem{xiong2020unsupervised}
W.~Xiong, D.~Liu, X.~Shen, C.~Fang, and J.~Luo, ``Unsupervised real-world
  low-light image enhancement with decoupled networks,'' \emph{arXiv preprint
  arXiv:2005.02818}, 2020.

\bibitem{zhang2019kindling}
Y.~Zhang, J.~Zhang, and X.~Guo, ``Kindling the darkness: A practical low-light
  image enhancer,'' in \emph{Proceedings of the 27th ACM International
  Conference on Multimedia}, 2019, pp. 1632--1640.

\bibitem{zhang2020self}
Y.~Zhang, X.~Di, B.~Zhang, and C.~Wang, ``Self-supervised image enhancement
  network: Training with low light images only,'' \emph{arXiv}, pp.
  arXiv--2002, 2020.

\bibitem{zhou2016evaluating}
H.~Zhou, T.~Sattler, and D.~W. Jacobs, ``Evaluating local features for
  day-night matching,'' in \emph{European Conference on Computer Vision}.\hskip
  1em plus 0.5em minus 0.4em\relax Springer, 2016, pp. 724--736.

\bibitem{elik2011ContextualAV}
T.~Çelik and T.~Tjahjadi, ``Contextual and variational contrast enhancement,''
  \emph{IEEE Transactions on Image Processing}, vol.~20, pp. 3431--3441, 2011.

\end{thebibliography}

\end{document}